# A Developmental Neuro-Robotics Approach for Boosting the Recognition of Handwritten Digits

Alessandro Di Nuovo
Sheffield Robotics, Department of Computing
Sheffield Hallam University
Sheffield, S1 1WB, United Kingdom
ORCID: 0000-0003-2677-2650

*Abstract*— **Developmental psychology and neuroimaging research identified a close link between numbers and fingers, which can boost the initial number knowledge in children. Recent evidence shows that a simulation of the children's embodied strategies can improve the machine intelligence too. This article explores the application of embodied strategies to convolutional neural network models in the context of developmental neuro-robotics, where the training information is likely to be gradually acquired while operating rather than being abundant and fully available as the classical machine learning scenarios. The experimental analyses show that the proprioceptive information from the robot fingers can improve network accuracy in the recognition of handwritten Arabic digits when training examples and epochs are few. This result is comparable to brain imaging and longitudinal studies with young children. In conclusion, these findings also support the relevance of the embodiment in the case of artificial agents' training and show a possible way for the humanization of the learning process, where the robotic body can express the internal processes of artificial intelligence making it more understandable for humans.**

*Keywords*— ***Neurorobotics; Cognitive Science; Computer Vision and Pattern Recognition; Machine Intelligence; Number Cognition.***

## I. Introduction

The embodied cognition theory affirms that human cognitive skills can be extended through bodily experiences, such as manipulatives, gestures, and movements [1]–[3]. Number processing is particularly valuable because it can provide a window into the neuronal mechanisms of high-level brain functions [4]. Numbers constitute the building blocks of mathematics, a language of the human mind that can express the fundamental workings of the physical world and make the universe intelligible [5]. Therefore, understanding how the artificial sensorimotor system embodies numerical processes can also help to answer the wider question of how bodily (real or artificial) systems support and scaffold the development of abstract cognitive skills [6]. Within the embodied mathematics framework, fingers are spontaneous tools that play a crucial role in developing number cognition until a level of basic arithmetic is achieved, for details see recent reviews [7], [8].

Embracing this theory, Developmental Neuro-Robotics aims at designing "robots whose control has been modelled after some aspect of the brain. Since the brain is so closely coupled to the body and situated in the environment, Neurorobots can be a powerful tool for studying neural function in a holistic fashion." [9], [10] The application of embodied theory in artificial agents is among the motivations for designing new robotic platforms for research to resemble the shape of a human body, known as "humanoids", e.g. ASIMO [11], and in particular that of a child, notably iCub [12] and NAO [13]. One of the postulates of this approach is that the humanization of the learning process can help to make artificial intelligence more understandable for humans and may increase the acceptance of robots in social environments [14]. Neurorobotics is still making its first steps, but it has already been successfully applied in the modelling of embodied word learning as well as in the development of perceptual, social, language and numerical cognition [15], [16], and recently extended as far as the simulation of embodied motor and spatial imagery [17]–[19]. The neurorobotics approach has also been used to simulate neuropsychological disorders and test possible rehabilitation procedures [20].

A recent study by Di Nuovo and McClelland [21] confirmed an increased efficiency in the recognition of spoken digits by using the sensory-motor information from an artificial humanoid body, i.e. the child-robot iCub, which is one of the few platforms that has fully functional five-fingered hands [22]. The results of the simulated training shown several similarities with studies in developmental psychology, such as a quicker creation of a uniform number line.

In this article, we further progress the development of symbolic numerical reasoning in humanoid robots by modelling the recognition of handwritten Arabic digits. Our approach mimics the developmental plasticity of the human brain, where new abilities are built upon the previous ones. Indeed, we first re-enact the previous training by training a single layer to associate fixed finger representations, i.e. motor positions to open/close the robot's fingers, to the number classes from 0 to 9. Then, the pre-trained layer is integrated into a convolutional classifier with new layers to perform the recognition of the handwritten digits by building upon the previously learned association. This between the Arabic digits and finger activations was observed in neuroscientific studies, e.g. [23].

The objective is to demonstrate a higher efficiency of machine learning by mimicking how children behave while developing numerical cognition. Meanwhile, we also aim at validating an approach for the humanization of artificial training strategies that can make machine learning more understandable for humans. This will help to reduce the scepticism against the deep learning approaches which are currently considered black boxes. For instance, human teachers may simply open and close the robot's fingers to instruct the robot or correct the representation in case of error.

This work was supported by the EPSRC grant EP/P030033/1 (NUMBERS).

Furthermore, we aim at providing useful insights on biologically inspired strategies that can improve machine learning performance in the context of applied robotics, where the training information is likely to be gradually acquired while operating rather than being abundant and immediately available as in the majority of machine learning scenarios.

## II. RELATED WORK

The simulation of embodied mathematical learning in artificial learning agents was usually focused on testing developmental psychology and neuroscientific theories [24]. For example, inspired by the earlier work by Alibali and Di Russo[25], Ruciński et al. [26] presented a model in which pointing gestures significantly improve the counting accuracy of the humanoid robot iCub. Recently, Pecyna et al. [27] presented neuro-robotics models with a deep artificial neural network capable of number estimation via finger counting. The experimental studies showed an improvement in training speed of number estimation when the network is also producing finger counting in parallel with the estimated number. Di Nuovo et al. [28]–[30] investigated artificial models for learning digit recognition (visual) and number words (auditory) with finger counting (motor), to explore whether finger counting and its association with number words or digits could serve to bootstrap number cognition. Results of the various robotic experiments show that learning number word sequences together with finger sequencing speeds up the building of the neural network's internal representations resulting in qualitatively better patterns of the similarity between numbers. The internal representations of finger configurations can represent the ideal basis for the building of an embodied number representation in the robot, something in line with embodied and grounded cognition approaches to the study of mathematical cognitive processes. Similarly to young children, the use of finger counting and verbal counting strategies, the developmental neuro-robotic model develops internal representations that subsequently sustain the robot's learning of the basic arithmetic operation of addition [28]. Successively, Di Nuovo et al. [31] presented a deep learning approach with superior learning efficiency. The new model was validated in a simulation of the embodied learning behaviour of bi-cultural children, using different finger counting habits to support their number learning. Recently, Di Nuovo [32] presented a new "shallow" embodied model for number cognition, which incorporates a neural link observed in neuroscientific studies [23], providing preliminary information on the effectiveness of the embodied approach in the recognition of synthetic handwritten digits. The results show how the robot fingers are an embodied representation of the numerosity magnitude that is the ideal computational representation for artificial mathematical abilities [33]. Moving further to arithmetic, Di Nuovo [34] investigates a Long Short-Term Memory (LSTM) architecture for modelling the addition operation of handwritten digits using an embodied approach. The results confirm an improved accuracy in performing the simultaneous recognition and addition of the digits, also showing an odd number of split-five errors in line with what has been observed in studies with humans [35].

All these studies demonstrated the value of the developmental neurorobotics approach to simulate aspects of numerical cognition in artificial cognitive systems. However, at the best of our knowledge, the study presented in [21] is the only one that demonstrated the effectiveness of the approach in a real task, i.e. recognizing the digits of the Google Tensorflow Speech commands dataset, while maintaining a plausible setting in the context of early cognitive development [21].

In this article, we present a further step in the developmental of numerical cognition in robots, i.e. the recognition of written number symbols. This will also allow demonstrating the potential applicability of the developmental neurorobotics approach to a popular machine learning benchmark database: the MNIST database of handwritten digits [36].

## III. MATERIALS AND METHODS

### A. The MNIST Database of Handwritten Digits

To provide a numerical challenge to our models, we used a very popular and publicly available benchmark in machine vision: the MNIST database of handwritten digits [37]. The database contains a total of 70,000 images, sized 28-by-28 (784 pixels), of handwritten digits divided into a training set of 60,000 examples, and a test set of 10,000 examples. This benchmark database is best suited for testing models on real-world data whilst spending minimal effort for pre-processing and formatting. Pixel values are normalized in the [0,1] range.

### B. Simulated embodied representations

Pictures of the iCub finger representations are in Figure 1, which shows the right hand. The iCub provides motor proprioception (joint angles) of the fingers' motors, for a total of 7 degrees of freedom (DoF) for each hand as follows: 2 DoF for the thumb, index, and middle fingers, and one for controlling both ring and pinky fingers, which are "glued together". Note that the hardware limitation is also common in human beings, who often can't freely move these two fingers independently [38], indeed the finger configurations of each hand are replicating the American Sign Language number representation from 1 to 5. Representations with the left hand are specular, and they are used in addition to the fully open right hand to represent numbers from 6 (5+1) to 9 (5+4).

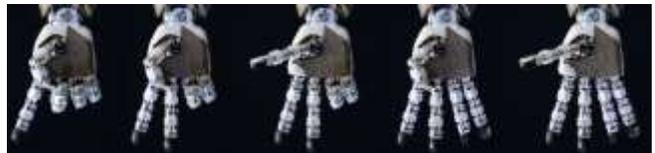

*Fig. 1*. The number representations with the iCub right-hand fingers. From one, two, three, four and five. Numbers from six to nine are represented with two hands, with the right hand fully open. In practice, the embodied inputs are the joint angles from the fingers' motor encoders.

The iCub robot encoder values for the finger representations indicate the number magnitude by the number of open fingers, though the numbers 3 and 4, as well as 8 and 9, involve only partially overlapping sets of fingers. To overcome the possible distortion by unbalanced representations because of the physical limitation, we duplicated the contribution of the motors that control the last two fingers; therefore, we have 16 inputs for the motor module. Encoder values are normalized in the [0,1] range. The actual numerical values of the iCub fingers representations can be found in [30] and in the GitHub repository (link is at the end of the article).

### C. Embodied Deep Learning architecture for recognising handwritten Arabic digit images

The artificial neural network architecture is based on a classical Convolutional Neural Network (CNN) classifier, the LeNet-5, which was proposed by LeCun et al. [39] for classifying handwritten digit images. This is a relatively simple but effective deep architecture that includes several layers that characterize the success of this approach: a sequence of 2 two-dimensional convolutional layers, each followed by Average Pooling layers, two densely connected ReLU layers and, finally, an output layer named "Classification_Layer" with a Softmax transfer function. Dropout and batch normalisation layers are inserted after each hidden layer to reduce overfitting and improve generalization performance.

A third model is also considered in our experiments. This was inspired by the Google Inception, which includes "auxiliary" classifiers to prevent the middle part of the network from "dying out" because of the limitations of propagating the error through the many layers of deep CNN [40]. The inception-like model doesn't include a pre-training.

The three models considered in our experimental comparisons are schematised in Figure 2.

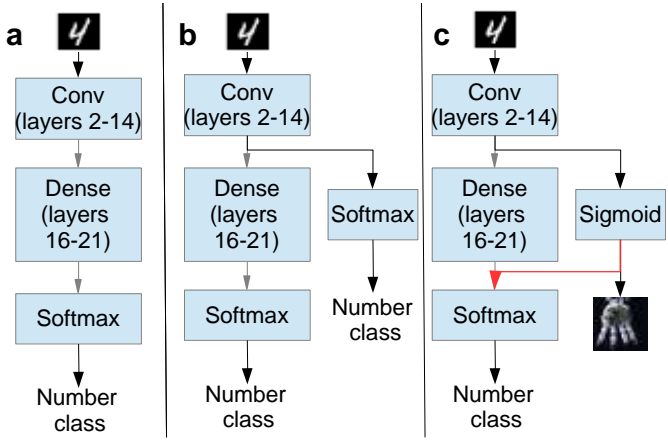

*Fig. 2*. The three models compared in our experiments: **(a) the baseline**, i.e. LeNet-5; **(b) the Inception-like** with the auxiliary *softmax* classifier; **(c) the embodied model** with the auxiliary output that is trained to predict robot's fingers configurations. Predicted fingers configurations are used to augment the final output layer (*softmax*). The weights of the link (in red) between the auxiliary layer (sigmoid) and the final output are pre-trained to simulate the previous learning [21].

The details of the general architecture with all the layers considered is summarised in Table I. The rows report the type, the size of the output, input and output links, the number of trainable parameters, the arguments and the initialization function for each layer. The baseline model (LeNet-5) includes all the layers, except the 15 (the auxiliary output - "Aux Out"), which is part of the Embodied model and the Inception-like. The pre-trained layer in the Embodied model is the final Softmax classifier (22). Out1 is the number class (one-hot representation), Out2 is the Robot hand encoders (fingers). The inception-like model the embodied layer (15) is not linked to the classifier (22) and he Out2 is the same one-hot representation of the number class.

The embodied model is realized by including an additional densely connected layer, the Auxiliary Output (n.15), which is connected and receives the input from the last Average Pooling layer. The embodied model is trained with a two-stage approach that simulated the developmental stages of learning: first, the final layer (n.22) is trained to associate embodied inputs to the number classes. Then, the remaining layers are connected, and the full model is tuned to classify the handwritten digits. Operatively, the auxiliary output layer n. 15 is inserted in the baseline model and it is fully connected to the output layer 22. The weights of this connection are those pre-trained in the first stage, therefore it simulates the neural link observed in neuroscientific studies [23].

TABLE I. A SUMMARY OF THE GENERAL CNN ARCHITECTURE.

| Layer | Type | Output Shape | In | Out | N. Param | Arguments |
|---|---|---|---|---|---|---|
| 1 | Inputs | 28x28 | Digits | 2 | | Range=[0,1] |
| 2 | Conv2D | 28x28 | 1 | 3 | 60 | filters=6, size=3x3; |
| 3 | Avg Pooling | 14x14 | 2 | 4 | | size=3x3; stride=2x2 |
| 4 | BatchNorm | 14x14 | 3 | 5 | 24 | |
| 5 | Dropout | 14x14 | 4 | 6 | | prob=0.2 |
| 6 | Conv2D | 14x14 | 5 | 7 | 880 | filters=16, size=3x3; |
| 7 | Avg Pooling | 7x7 | 6 | 8 | | size=3x3; stride=2x2 |
| 8 | BatchNorm | 7x7 | 7 | 9 | 64 | |
| 9 | Dropout | 7x7 | 8 | 10 | | prob=0.2 |
| 14 | Flatten | 784 | 13 | 15&16 | | |
| 15 | Dense (Aux Out) | 16 | 14 | Out2& 22 | | f=Sigmoid |
| 16 | Dense | 120 | 14 | 17 | 94200 | f=ReLU |
| 17 | BatchNorm | 120 | 16 | 18 | 480 | |
| 18 | Dropout | 120 | 17 | 18 | | prob=0.5 |
| 19 | Dense | 84 | 18 | 19 | 10164 | f=ReLU |
| 20 | BatchNorm | 84 | 19 | 20 | 336 | |
| 21 | Dropout | 84 | 20 | 21 | | prob=0.5 |
| 22 | Dense (Classifier) | 10 | 15&21 | Out1 | 850 / 1010 | f=Softmax |

### A. Convolutional Neural Network Implementation

The models were implemented, trained and tested using python and Keras 2.2.0 [41] high-level APIs running on top of TensorFlow 1.8.0 [42]. A link to the open-source repository is given at the end of this article. Greater detail on the APIs can be found in the documentation of these tools available from the respective websites [41], [42].

The details of the LeNet-5 architecture can be found in the original article [39]. However, we updated the architecture by adding two types of layers, which were proved to be beneficial for the generalisation. The Dropout layer, which operates by randomly dropping a fraction of input at each update at training time. Dropout layers help to prevent overfitting. Dropout rates were 0.2 for the convolutional layers, while they were 0.5 for the densely connected layers. Like all the other parameters, these rates were shown to be optimal by previous extensive analysis, also including the MNIST dataset [43], [44]. The Batch Normalization layer, which scales the output of the previous layer by standardizing the activations of each input variable per mini-batch. This has the effect of inducing a more predictive and stable behaviour of the gradients, which allows faster training [45].

### B. Algorithms for training the networks

For the training, we selected the most famous adaptive learning method, based on stochastic gradient descent, for training the models: The Adaptive Moment Estimation algorithm (Adam) [46]. The training was executed in mini-batches of 32 examples (128 in the case of the full database). The use of mini-batches proved to improve the generalization of the network, i.e. the accuracy in the test set. As recommended, we left the parameters of this optimizer at their default values, which follow those provided in the original publications cited. Adam is widely used in the field of deep learning because it is fast and achieves good results. Adam is a gradient-based method that maintains per-parameter ($\theta$) learning rates:

$$\theta(t+1) = \theta(t) - \frac{\eta}{\sqrt{\hat{v}(\theta,t)}} \cdot \frac{\partial L}{\partial \theta}(t)$$

Where $\frac{\partial L}{\partial \theta}(t)$ is the gradient of the loss function $L(t)$ at epoch $t$, $\eta$ is the learning rate, which, in our experiments, has been set as *0.001* and $\gamma$ is 0.9. Adam makes use of a moving average of the squared gradient $\hat{v}(\theta,t)$:

$$\hat{v}(\theta,t) = \gamma \cdot \hat{v}(\theta,t-1) + (1-\gamma) \cdot \left(\frac{\partial L}{\partial \theta}(t)\right)^2$$

while it keeps an exponentially decaying average of past gradients $\hat{m}(\theta,t)$, similar to the momentum. Adam's parameter update is given by:

$$\theta(t+1) = \theta(t) - \eta \frac{\hat{m}(\theta,t)}{\sqrt{\hat{v}(\theta,t)}}$$

Specifically, $\hat{v}(\theta,t)$ and $\hat{m}(\theta,t)$ are calculated using the parameters $\beta_1$ and $\beta_2$ to control the decay rates of these moving averages:

$$\hat{m}(\theta,t+1) = \frac{m(\theta,t)}{1-\beta_1^t}$$

where $m(\theta,t) = \beta_1 \cdot m(\theta,t-1) + (1-\beta_1) \cdot \frac{\partial L}{\partial \theta}(t)$

$$\hat{v}(\theta,t+1) = \frac{v(\theta,t)}{1-\beta_2^t}$$

where $v(\theta,t) = \beta_2 \cdot v(\theta,t-1) + (1-\beta_2) \cdot \left(\frac{\partial L}{\partial \theta}(t)\right)^2$

Note that $\beta_1^t$ and $\beta_2^t$ denote the parameters $\beta_1$ and $\beta_2$ to the power of $t$. Suggested default settings are $\eta = 0.001$, $\beta_1 = 0.9$ and $\beta_2 = 0.999$. These values are used in our experiments.

*1) Loss function*

The Loss function $L(t)$ is the Cross-entropy function, which computes the performance given by network outputs and targets in such a way that extremely inaccurate outputs are heavily penalized, while a very small penalty is given to almost correct classifications.

The calculation of the Cross-entropy depends on the task: Categorical $H_C$ when classifying into the number classes; Binary $H_B$ predicting the embodied representations.

In the case of classification, the output $\boldsymbol{p}$ is a categorical vector of N probabilities that represent the likelihood of each of the N classes with $\sum \boldsymbol{p} = 1$, while $\widetilde{\boldsymbol{y}}$ is a one-hot encoded vector (1 for the target class, 0 for the rest). The Categorical cross-entropy $H_C$ is calculated as the average of the cross-entropy of each pair of output-target elements (classes):

$$H_C = \frac{1}{N}\sum_{i=1}^{N} -\widetilde{y}_i \cdot \log(p_i)$$

When the target is the embodied representation, the output is a vector $\boldsymbol{z}$ of $K$ independent elements. The cross-entropy can be calculated considering 2 binary classes: one corresponds to the target value, zero otherwise. In this case, the loss function is calculated using the binary cross-entropy expression:

$$H_B = \frac{1}{K}\sum_{i=1}^{K} -\widetilde{y}_i \cdot \log(z_i) - (1-\widetilde{y}_i) \cdot \log(1-z_i)$$

In the training phase of the embodied model, the loss is the weighted sum of the losses for the two outputs (Categorical and Binary Cross-entropy), weighted respectively: 1 for the classification loss; a varying value between 1 and 0.1, depending on the number of examples, for the embodied loss.

### C. Statistical Analysis

We used the Student t-test to calculate the p-value and confirm the statistical significance of the comparisons. To evaluate the effect size of the differences, we calculated Cohen's d [47]. Cohen suggested that d=0.2 be considered a "small" (trivial) effect size, 0.5 represents a "medium" effect size and 0.8 a "large" effect size [47].

## IV. EXPERIMENTAL RESULTS

This section presents the experimental results of the models obtained by training and testing the CNN architecture with the MNIST handwritten digits database. The analysis simulates a gradual course of education typical for the children, we have investigated the models' performance of varying size of training examples: 256, 512, 1024, 3200, 6400. These can also be considered as different levels of "practice". We also considered the full 60000 for information and comparability with the other experiments in the machine learning literature. The embodied models are compared against the standard models built with visual inputs only, which constitute the baseline for the comparison. The results of the inception-like model are presented for additional comparison with a state-of-the-art approach.

Models were trained for 20 epochs as in [39]. Each model was trained and tested 21 times. The Figures and Tables in this section report the average accuracy and the standard deviation of the 21 repetitions, with a particular focus on the development of the learning, i.e. how the performance progresses from the first epoch until the last one with an increasing number of training examples. In the following, we present an analysis of the models' performance, i.e. accuracy of recognition, on the whole database (examples used for training plus testing) and on the testing set only.

### A. Analysis of the results on the whole database

In several possible applications, robots operate in a restricted environment in which they may be required to recognise the same digits that are presented for the training. For instance, this is the case of personal assistants that will serve a few users, who are likely to present very similar digits for training and recognition. In these cases, it is essential to quickly achieve the best accuracy in both the training and testing set. Here we analyse the performance on the whole database, i.e. training and testing. For example, in the smallest size considered, the models are trained with 256 examples, then we present the results on the prediction of 10256 digits, i.e. 256 training plus 10000 testing examples.

TABLE II. ACCURACY RATES FOR VARYING TRAINING EXAMPLE SIZES (THE WHOLE DATABASE).

| Examples | Embodied | | Inception-like | | | Baseline | | |
|---|---|---|---|---|---|---|---|---|
| | avg | stdev | avg | stdev | d | avg | stdev | d |
| After 1 epoch | | | | | | | | |
| 256 | 0.272 | 0.044 | 0.250 | 0.030 | -0.534 | 0.271 | 0.050 | -0.030 |
| 512 | **0.478** | 0.050 | 0.433 | 0.041 | -0.889 | 0.428 | 0.053 | -0.880 |
| 1024 | **0.693** | 0.024 | 0.606 | 0.037 | -2.582 | 0.586 | 0.033 | -3.413 |
| 3200 | **0.824** | 0.012 | 0.731 | 0.017 | -5.811 | 0.733 | 0.014 | -6.335 |
| 6400 | **0.856** | 0.008 | 0.750 | 0.010 | -10.451 | 0.756 | 0.012 | -8.797 |
| 60000 | **0.892** | 0.006 | 0.823 | 0.020 | -4.346 | 0.817 | 0.009 | -8.845 |
| After 20 epochs | | | | | | | | |
| 256 | **0.845** | 0.013 | 0.830 | 0.024 | -0.740 | 0.818 | 0.026 | -1.182 |
| 512 | **0.906** | 0.009 | 0.891 | 0.016 | -1.054 | 0.885 | 0.013 | -1.763 |
| 1024 | **0.943** | 0.006 | 0.931 | 0.006 | -1.847 | 0.926 | 0.008 | -2.168 |
| 3200 | **0.969** | 0.003 | 0.959 | 0.003 | -2.729 | 0.961 | 0.002 | -2.600 |
| 6400 | **0.977** | 0.001 | 0.967 | 0.002 | -4.993 | 0.966 | 0.002 | -5.524 |
| 60000 | **0.986** | 0.001 | 0.980 | 0.001 | -6.329 | 0.980 | 0.001 | -7.800 |

Table II presents the Average accuracies (Avg) on the whole dataset, with standard deviations (StDev) and Cohen's d. Values in bold are significantly (p<0.05) better than the others.

Table II shows that the embodied model is significantly better than the baseline and the inception-like models in all case considered and with a large (>0.8) effect size. There is no statistical significance only in the first few epochs of the smallest sample size.

The full development of the training is presented in Figure 3, where the graph shows the accuracy rate of all the 20 epochs for three sample sizes. Figure 2 compares the baseline (yellow lines), the embodied model with the iCub robot fingers (blue lines) and the Inception-like model (red lines). A significant increase in the accuracy is visible in the first epochs and for all sample sizes, while the improvement is reduced but still significantly present in all the other cases. Figure 2 shows also that the accuracy, of both the baseline and the embodied model, reaches a maximum after around 10-15 epochs and there is no overfitting.

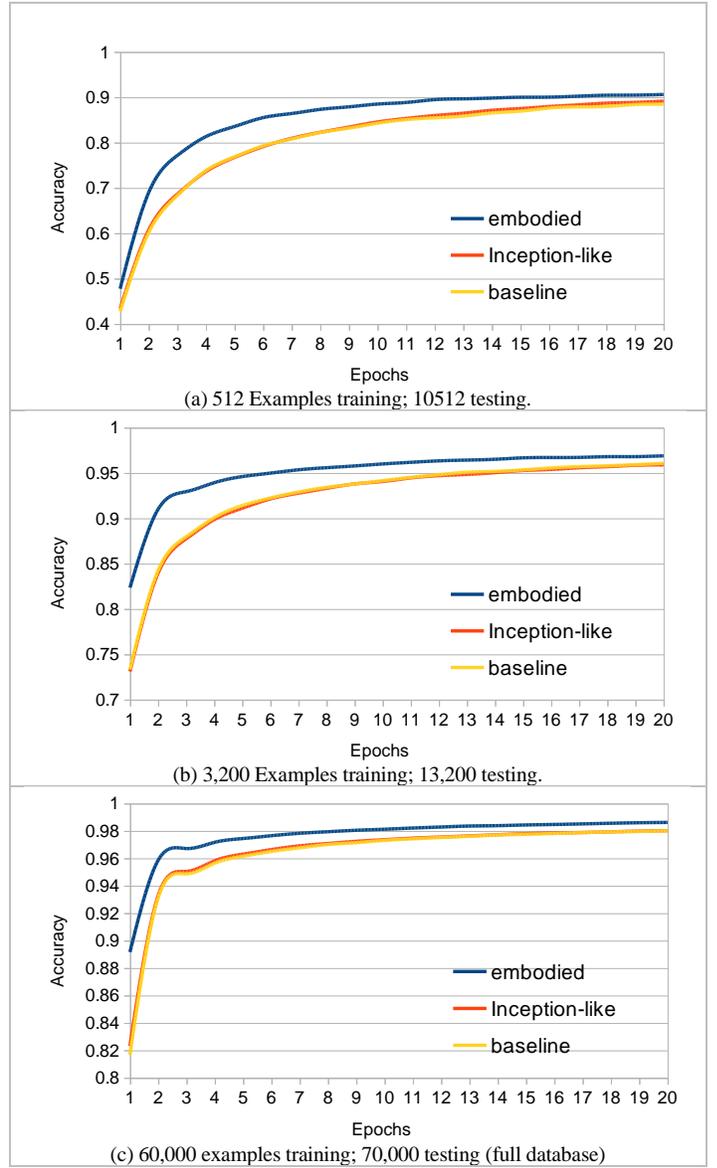

(a) 512 Examples training; 10512 testing.

(b) 3,200 Examples training; 13,200 testing.

(c) 60,000 examples training; 70,000 testing (full database)

*Fig. 3*. The accuracy rate of the models on the whole database (blue: robot/embodied; red: Inception-like; yellow: Baseline). a, small: 10,512 examples; b, medium: 13,200 examples; c, full database (70,000).

### B. Analysis of performance on the testing set

This section presents the usual analysis of the performance on the testing set as required for validation and for showing the generalisation abilities of the models.

Table III presents the Average accuracies (Avg) on the test set, with standard deviations (StDev) and Cohen's d. Values in bold are significantly (p<0.05) better than the others.

TABLE III. ACCURACY RATES FOR VARYING TRAINING EXAMPLE SIZES (TESTING SET).

| Examples | Embodied | | Inception-like | | | Baseline | | |
|---|---|---|---|---|---|---|---|---|
| | avg | stdev | avg | stdev | d | avg | stdev | d |
| After 1 epoch | | | | | | | | |
| 256 | 0.267 | 0.045 | 0.251 | 0.031 | -0.358 | 0.266 | 0.051 | -0.008 |
| 512 | **0.494** | 0.051 | 0.447 | 0.043 | -0.891 | 0.438 | 0.056 | -0.943 |
| 1024 | **0.722** | 0.024 | 0.645 | 0.039 | -2.181 | 0.626 | 0.035 | -2.887 |
| 3200 | **0.882** | 0.011 | 0.834 | 0.017 | -3.000 | 0.840 | 0.015 | -2.875 |
| 6400 | **0.929** | 0.007 | 0.896 | 0.013 | -2.974 | 0.899 | 0.011 | -2.967 |
| 60000 | **0.971** | 0.003 | 0.963 | 0.005 | -1.671 | 0.964 | 0.004 | -1.865 |
| After 20 epochs | | | | | | | | |
| 256 | **0.844** | 0.014 | 0.831 | 0.024 | -0.596 | 0.820 | 0.027 | -1.038 |
| 512 | **0.905** | 0.009 | 0.893 | 0.016 | -0.871 | 0.887 | 0.013 | -1.449 |
| 1024 | **0.940** | 0.007 | 0.935 | 0.006 | -0.772 | 0.929 | 0.007 | -1.408 |
| 3200 | 0.968 | 0.004 | 0.966 | 0.003 | -0.271 | 0.968 | 0.002 | 0.085 |
| 6400 | 0.977 | 0.002 | 0.977 | 0.002 | -0.125 | 0.976 | 0.002 | -0.406 |
| 60000 | 0.991 | 0.001 | 0.990 | 0.001 | -0.432 | 0.991 | 0.001 | -0.307 |

Table III confirms the initial advantage of the embodied model over the others. Though, the advantage is lost with the medium-large sizes of training data, where there is no statistically significant difference among the three models. However, after 20 epochs, the recognition rate of the embodied model is usually higher with a medium effect size.

Figure 4 shows the development of the accuracy on the training set on three examples with different sample sizes. The graphs show the accuracy rate of all the 20 epochs for the testing set only. Figure 3 presents the baseline (yellow lines), the embodied model with the iCub robot fingers (blue lines) and the Inception-like model (red lines). A significant increase in the accuracy is clearly visible in the first epochs, except in the case of 256 samples for training where it took 4 epochs before the increase became significant. The improvement is reduced but still significantly present usually until around 10-15 epochs, where the embodied model converges. Figure 3 shows also that there is no overfitting.

*C. Discussion*

Summarising the results of the analyses, we see that the embodied approach improves the fitting with the training data without affecting the generalisation of the training. Moreover, when the training examples are few, the embodied approach is also beneficial to improve the generalisation in the test set.

The experiments show that extending the training with additional output, e.g. the inception-like, doesn't seem beneficial in the cases considered. There is an increased efficiency only if the training is augmented by representations that indicate the number magnitude, which can help to disambiguate and, therefore, speed-up the training as seen for the spoken digits [21]. We suggest the use of a finger-based representation because this can be naturally created by a humanoid robot via interaction with a human teacher, who can open and close the fingers of the robot to provide the additional information.

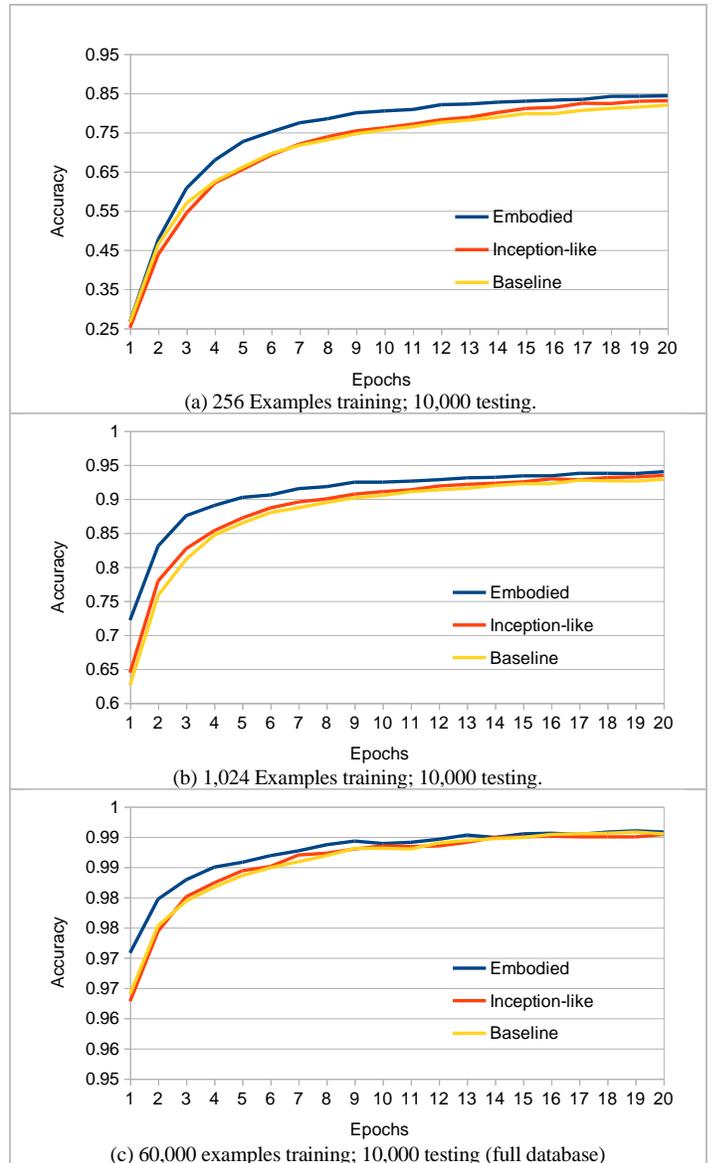

*Fig. 4.* The accuracy rate of the models on the test set (blue: robot/embodied; red: Inception-like; yellow: Baseline). a, small: 256 training examples; b, medium: 1,024 training examples; c, 60,000 training examples.

V. CONCLUSION

Recent studies in developmental psychology and cognitive neuroscience demonstrated a pivotal role of fingers in developing number cognition. Inspired by these studies, this article presented an investigation on the development of handwritten digit recognition when convolutional neural networks are embodied in the humanoid robot iCub's and its proprioceptive information can be used to augment the training. Indeed, we created a model that integrates the previous experience that links finger configurations to number classes. The model is then trained to build upon this link to predict the finger configurations as an additional output and, at the same time, as an input for the classifier.

The experiments with the embodied approach show that the robot's fingers can boost the learning, especially when the training examples are numerically limited. This is a common scenario in applied robotics, where it is likely that robots will learn with only a small amount of data, acquired through interaction.

From the machine learning point of view, the embodied strategy could be also seen as a bio-inspired extension to the "auxiliary" classifiers that were introduced in the Google Inception network to prevent the middle part of the network from "dying out" because of the limitations of backpropagation algorithms in propagating the error through the many layers of deep CNN [40]. Indeed, the experimental comparison with an inception-like model shows the higher efficiency of the embodied approach in the case of limited training set sizes.

The analysis of the results shows similarities with the transition from early to mature mathematical cognition observed in longitudinal studies with children, who initially perform better when they can use fingers, but, after they grow in experience, gradually abandon finger representations without affecting the accuracy [48]. We believe that these findings can confirm the importance of the body for developing number knowledge also in artificial neural network models and show the benefits of the embodied cognition ideas also in the machine learning context.

Future studies will focus on multimodal models that can support the investigation on the interconnections between spoken and written digits. Furthermore, realistic and in-the-wild applications will be investigated, for instance, interactive learning from a human teacher with the use of vision (camera inputs) to recognise robot's and human's fingers [49].

ACKNOWLEDGMENT

The author wishes to thank prof. James McClelland for the valuable comments and discussion about this work. The author is grateful to the NVIDIA Corporation for the donation of a Tesla K40 and a GeForce TITAN X that have been used for the experiments.

CODE AVAILABILITY

The data and source code for the models presented in this paper can be found in the GitHub repository: https://github.com/EPSRC-NUMBERS/EmbodiedCNN-MNIST.